\documentstyle[ijcai89,named]{article}

\newcounter{cex}
\newcounter{cexi}
\newcounter{cexii}

\def\ex{\refstepcounter{cex} \setcounter{cexi}{0} \item}
\def\exi{\refstepcounter{cexi} \setcounter{cexii}{0} \item}

\newenvironment{el}
{
 \begin{list}{(\arabic{cex})}{
    \topsep = 1.2ex
    \itemsep = 1.0ex
    \parsep = 0pt
    \rightmargin = 1.0em
    \leftmargin = 3.7em
    \labelwidth = 2.5em
    \labelsep = 1.5em
 }}{ \end{list} }

\newenvironment{eli}
{
 \begin{list}{\alph{cexi}.}{
    \itemsep = 0pt
    \parsep = 0pt
    \rightmargin = 0em
    \leftmargin = 1.4em
    \labelwidth = 1.0em
    \labelsep = 1.0em
 }}{ \end{list} }

\newlength{\badstar}

\title{\bf A Study of the Context(s) in a Specific Type of Texts:
Car Accident Reports}

\author{
Dominique Estival \\
ISSCO, Universit\'{e} de Gen\`{e}ve\\
54 rte des Acacias, CH-1227 Gen\`{e}ve\\
$<$estival@divsun.unige.ch$>$\\
+41-22-705-71-16
\And
Fran\c{c}oise Gayral\\
LIPN, Universit\'{e} Paris-Nord\\
Av. J.-B. Cl\'{e}ment, F-93430 Villetaneuse\\
$<$fg@lipn.univ-paris13.fr$>$\\
+33-1-49-40-36-25
}

\begin{document}

\maketitle

\begin{abstract}

This paper addresses the issue of defining context, and more specifically
the different contexts needed for understanding a particular type of texts.
The corpus chosen is homogeneous and allows us to determine characteristic
properties of the texts from which certain inferences can be drawn by the
reader.  These characteristic properties come from the real world domain
(K-context), the type of events the texts describe (F-context) and the genre
of the texts (E-context).  Together, these three contexts provide elements
for the resolution of anaphoric expressions and for several types of
disambiguation.  We show in particular that the argumentation aspect of
these texts is an essential part of the context and explains some of the
inferences that can be drawn.

\end{abstract}

\section{Introduction}

We must first emphasize that our objectives in this paper are not general
considerations about context or the theory of context, but that they are
guided by the particular goals of a specific project.  The work reported
here is the result of a study done within a larger project on the
``Semantics of Natural Languages'', viewed from the fields of Artificial
Intelligence and Computational Linguistics, in which we are treating a
corpus of real texts.  The corpus consists of a number of insurance claim
reports for car accidents.\footnote{
The project {\em S\'{e}mantiques des Langues Naturelles\/}, sponsored
by the French Minist\`{e}re de la Recherche and the CNRS, involves a
number of research centers and university laboratories \cite{Gsem90a},
\cite{Estival/Gayral94}.
The texts were provided by the French insurance company MAIF, after being
made anonymous.  They have been translated into English by D.E.}
This corpus offers a number of advantages, the main one being its unity,
with respect to a) the domain involved (which is relatively circumscribed)
and b) the conditions of enunciation (which are almost always the same).
Indeed, the texts are written after an accident and their
quasi-institutional nature imposes a number of constraints on their
production and, in a symmetric way, on their interpretation.
This allows us to focus on the characteristic properties of the type of the
texts, and to draw inferences from them.

We can then better define the notion of context and answer the following
questions, which are those we address more specifically in this paper:

$\bullet$ In which way does context affect interpretation of NL utterances
and texts?

$\bullet$ Which aspects of context or which contexts result in refined,
more general, and different interpretations of NL?

$\bullet$ What is context?

$\bullet$ How many contexts are there?

\section{The Contexts of the Corpus}

All the texts are written in similar circumstances and belong to a
culturally well-defined genre which both writer and reader are perfectly
aware of when writing or reading one of them.  An insurance claim report for
a car accident is not a newspaper story, nor a letter to a friend narrating
the accident, but an almost institutionalized document which obeys certains
constraints concerning its content as well as its form.  As emphasized by
Rastier in \cite{Rastier/Cavazza/Abeille94}, genre constraints play a major
role in the interpretation process.  In order to describe these constraints,
we distinguish three types of contexts, which we refer to as:

\begin{itemize}
\item {\bf K-context} (knowledge context):
This is what is usually meant by ``context'' and it refers to the
extra-linguistic knowledge for the particular domain of the texts.
In this case, our K-context is that of the world of road traffic, and
it concerns vehicles, vehicle motions, traffic rules, the usual
behavior of drivers and pedestrians, as well as their expectations,
and also some elements of ``naive'' geometry.

\item {\bf F-context} (factual context):
For this type of texts, there are two factual constraints bearing on
the {\bf content} of the text:

--  the text is a narration in which an accident takes place;

--  the text involves at least two participants, generally two
vehicles, one of which is the author's.

\item {\bf E-context} (context of the enunciation):
The conditions of enunciation (the {\bf discourse constraints})
for those texts are:

--  an imposed format:
Before writing their text, the authors must check some boxes on the
insurance claim form.  These boxes are labelled with ready-made
expressions and phrases which influence the vocabulary that is
then used in the running text (terms such as {\em vehicle\/}, or the
use of the labels ``A'' and ``B'' for each of the protagonists).
On the insurance claim form, the space in which the author can write
this text is pre-defined.  It is rather small and thus the text must
be rather short, at most one paragraph.

--  known addressee:
The recipient of the text is known, it is the insurance company.
Thus the argumentative aspect of the text is also known in advance:
the authors of the texts will try to lessen their responsibility.

\end{itemize}

We will show how these three context types are used by the authors to write
their texts and symmetrically by the readers to interpret them.  These two
symmetrical tasks are both composed of a factual and an argumentative part,
which coincide with the two goals we can define for an NLP approach to both
understanding and processing these texts.  One of these goals is the factual
analysis which is necessary to recreate the event:  ``What happened?  What
real world events concerning the motions of these vehicles or the scene
geometry actually occurred?''.\footnote{
We \cite{Estival/Gayral94} take an approach close to that of
\cite{Barwise88}, where the notation ``$P = C_{LC}(T,c)$'' denotes the
informative content P of the text T used in circumstances c, with the
language conventions LC shared by all the participants.}
The other is an argumentative analysis which takes into account the nature
and intent of the text in order to uncover the argumentative devices used by
the writer.

Although we can thus clearly delineate these two goals, they may not be so
neatly separated in practice and we find that, in real text processing, they
are intertwined in such a way that solving one level of analysis requires
elements from the other.  Similarly, although the division of context into
three context types is extremely useful and revealing, we sometimes have to
invoke more than one of them to treat some aspects of our texts.
Nevertheless, we organize the remainder of this paper along the way these
three context types can be used to describe the processes necessary for both
the production and interpretation of our texts.

\section{The three types of Contexts}

Whereas E-context and F-context are particular to this type of texts,
K-context is independent of the type of texts:  the knowledge involved will
be the same whether the text is an accident report or a newspaper article
and any text dealing with the road domain will invoke the same K-context.

\subsection{K-context}
\label{K-context}

K-context can be taken as the domain ontology for these texts.  It is
already well-established that domain ontology is necessary for natural
language understanding and that purely linguistic knowledge is not
sufficient.  In our texts, K-context includes knowledge about the Rules of
the Road, about driving, and the typical knowledge about the objects evoked
and their relations with each other, hierarchical or otherwise.  This
knowledge is shared by the writer ({\em W\/}) and the reader ({\em R\/}) and
it is used by the reader in a number of specific tasks, in particular to
solve anaphors and to make inferences.

\subsubsection{Rules of the Road}

In these texts, the vehicles are supposed to be moving in a space which is
regulated by the French traffic rules ({\em Code de la route\/}) and the
drivers are supposed to obey those rules.  For instance, we can see how
knowledge about stop-signs is used when we examine the reasoning which must be
made
by the reader of \ref{A1} to reconstruct the scene (two vehicles, with
their right blinker on, are stopped in front of the writer's at a
stop-sign).

\begin{el}
\ex {\em I was at a stop-sign with two cars in front of me turning to the right
towards Mours.  While the first car was going through this stop-sign I
performed my check to the left and started but I hit the second car which
hadn't yet gone through the stop-sign\/}  (A1)\label{A1}
\end{el}

First, the knowledge that ``Drivers must stop at a stop-sign.'' is useful to
infer that the interpretation for {\em I was at a stop-sign\/} here is
clearly {\em I was stopped at a stop-sign\/}, although the verb {\em to
be\/} in the past tense (an imperfect in the original text) followed by a
locative adverbial does not necessarily entail that its subject is stopped;
indeed, one can say {\em I was on the highway\/} without implying that one
was not moving.

Second, the reader uses the rule ``If X is at a stop-sign and X has switched
his right blinker on, X will turn right'' to interpret {\em with two cars in
front of me turning to the right\/} as ``The two cars were stopped and had
switched their right blinker on'' rather than ``The two cars were turning
right'', an interpretation which the present participle would allow.  This
interpretation requires some reasoning which is very difficult to automatize in
a
computer program.
Indeed, since {\em W\/} says that the two vehicles were turning right while
in fact they were stopped, the expression {\em turning right\/} proves to be
only an intention:  the cars were stopped but they were ``going to turn
right''.  If this intention had remained in the mind of the driver, it would
have been opaque to {\em W\/}, therefore an element expressing it must
have been visible and perceived by {\em W\/}:  the right blinker.

Third, the sentence {\em I performed my check to the left\/} is correctly
understood by {\em R\/} because {\em W\/} and {\em R\/} both know what
actions are expected at a stop-sign through the stereotyped knowledge or
{\bf script} \cite{Schank/Abelson77} for ``X being at a stop-sign'':  it
implies both that X will not stay at this stop-sign and that, in order to go
through it, X must check the road for safety, i.e.\ check that no vehicle is
coming.

Drivers also know what they must do in case of an accident:  they should
fill in a car accident report, sign it, get it signed by the other driver(s)
(and witnesses if any) involved, and send it to their insurance company.
In \ref{A3}, several linguistic elements can only be understood through
the knowledge of what is expected in that situation.

\begin{el}
\ex
{\em Heavy traffic on Bd Sebastopol.  I was driving between two lanes of
stopped cars when one of the cars on my left opened its right front door.
To avoid it, I swerved, which made me touch vehicle B with the rear of
my motorcycle, which made me fall.  Because of the heavy traffic that day,
we \underline{only} exchanged our insurance companies and names
\underline{which} explains why the report is only signed by me.\/}
(A3)\label{A3}
\end{el}

The second relative pronoun {\em which\/} refers not to the fact of having
exchanged insurance information and names, but to the fact that this is all
that happened, while much more would have been expected from the accident
script:  i.e.\ get the report form, check the appropriate boxes, make the
drawing, write the report, and sign it.
The word {\em only\/} ({\em juste\/} in the French text) signals that
something is missing, that there is a deviation with respect to the
situation expected from the K-context.  The relative pronoun thus refers to
what is actually missing from the reported scene, but which would have been
implied otherwise.

\subsubsection{Typicality}

Typicality can first be considered at the lexical level.  Indeed, any domain
ontology induces certain preferences for the interpretation of lexical items
which may have several meanings, and there are examples in our texts where
lexical typicality helps resolve polysemy (which may not even be noticed
by a human reader, but would cause problems in machine processing).

For example, in the phrase {\em je roulais\/} which occurs very often in our
texts, e.g.\ in \ref{A8}, the verb {\em rouler\/} must be interpreted as
``to travel by means of a vehicle with wheels'' rather than as ``to roll''
(as in {\em Paul roule dans le sable/ Paul rolls in the sand\/}) because the
K-context makes it improbable that a text describing a car accident would
talk about somebody rolling in the second sense.

\begin{el}
\ex
Je \underline{roulais} sur la partie droite de la chauss\'{e}e quand un
v\'{e}hicule arrivant en face dans le virage a \'{e}t\'{e}
compl\`{e}tement d\'{e}port\'{e}.  Serrant \`{a} droite au maximum, je n'ai
pu \'{e}viter la voiture qui arrivait \`{a} grande vitesse.

{\em I was driving on the right hand side of the road when a vehicle
arriving in front of me in the curve was completely thrown off course.
Keeping as close as possible to the right, I wasn't able to avoid the car
which was coming with great speed.}  (A8)\label{A8}
\end{el}

The notion of typicality (relative to a domain or a context) also concerns
the knowledge of which entities are considered typical in that context.
This is crucial for determining what the entities mentioned in a text are,
and can also be illustrated with \ref{A8}, where {\em R\/} must
determine that the two different expressions {\em a vehicle arriving in
front of me in the curve\/} and {\em the car which was coming with great
speed\/} are co-referent and thus that the text only involves two vehicles.

First, the two terms being used, {\em vehicle\/} and {\em car\/}, are
compatible, indeed a car is a particular type of vehicle.  This fact can be
extracted from a hierarchy of concepts which is part of K-context.
Second, in Western industrialized countries the most typical vehicle is a
car, so without any other indication a vehicle is inferred to be, typically
or by default, a car.\footnote{
Of course, typicality is a strongly cultural notion.  In China, for
instance, the most typical vehicle could be a bicycle and the word {\em
vehicle\/} might by default refer to a bicycle.  We assume here that the
determination of typicality in a language is mediated through the actual
hierarchy of concepts which is culturally defined.}
In \ref{A8}, since {\em W\/} first uses the word {\em vehicle\/} to
introduce an object, {\em R\/} infers that this vehicle is a car; this
default conclusion is confirmed by the next expression {\em the car which
was coming with great speed\/}.

The way {\em R\/} uses this {\bf Typicality Rule} can be illustrated with
the end of \ref{A3}.  As usual, the expression {\em I was driving\/}
implicitly introduces a vehicle, which is by default a typical one, i.e.\ a
car.  {\em R\/} thus starts building a representation for the scene with a car
as {\em W\/}'s vehicle until the expression {\em with the rear of my
motorcycle\/} forces him to reconsider his previous interpretation of {\em I
was driving\/} as {\em I was driving a car\/}.  Moreover, {\em R\/} must then
also reconsider the meaning of {\em between two lanes of stopped cars\/},
in particular the actual width of space associated with that expression.
The first interpretation induces a representation where there are three car
lanes and vehicle A is in the middle one (which is moving), while the second
interpretation leads to the correct spatial representation, where there are
only two car lanes and vehicle A is between them.

{}From {\em W\/}'s point of view, the Typicality Rule can be linked to
Grice's Maxim of Quantity:  ``Be as informative as possible''.  If the
vehicle in the scene is not a typical one, {\em W\/} should say so, or else
he would be hiding an important piece of information, useful for {\em R\/}
to interpret the text.  The mild strangeness of \ref{A3} can be
explained by this minor violation.

Associations between entities may also be more or less typical relative to a
particular context, and these typical associations helps resolve associative
anaphors and metonymies.  For instance, in \ref{A17}, knowing that there
are gas pumps in a gas station allows {\em R\/} to link {\em the pump\/} to
the gas station which is mentioned in the previous sentence.  Similarly, we
know that when a car door is smashed, someone must be responsible, which
explains the definite article in
\ref{A2bis}.

\begin{el}
\ex
\begin{eli}
\exi {\em I was entering (vehicle A) the lane into a gas station.
\underline{The pump} being out of order, I was backing up to leave when I
hit vehicle B which had also entered the same lane to get gas.\/}
(A17)\label{A17}

\exi {\em Having left my car to call a mechanic, I came back to find it with
the right back door bashed in with no note left by \underline{the} guilty
party.\/} (A2)\label{A2bis}
\end{eli}
\end{el}

Metonymy is a general linguistic device, i.e.\ part of LC, which is often
used to allow the identification (inter alia) of a container with its
content, and a common use of metonymy in our texts concerns the vehicle and
its driver.  Metonymy creates a unique discourse entity with properties
coming from the elements being identified.  A common use of metonymy in our
texts concerns the vehicle and its driver.  This metonymy allows
transference of properties either from the driver to the car, e.g.\
intentionality in\ref{1} and agentivity in \ref{A7}, or from the car to its
driver, as in \ref{bumper} where objects (here the bumper) belonging to the
vehicle are treated as belonging to the driver, or \ref{rolling} where the
property of ``rolling along'' (the literal meaning of the verb {\em
rouler\/}) is transferred to the driver.\footnote{
We can note that the use of this metonymy is not symmetrical:  the other
protagonist is aften only perceived through his car (i.e.\ car$->$driver),
while the writer sees his car as an extension of himself (i.e.\
driver$->$car).}

This use of metonymy follows the coercion of semantic types (see
\cite{Pustejovsky89b}) in a predictable way:  the properties being used to
make an entity of one type (e.g.\ {\em car\/}:  ``inanimate mechanical
object'') into an entity of another type (e.g.\ {\em driver\/}:  ``human
agent'') are extractible in a regular way from the predicate (e.g.\ {\em
squeeze\/}:  ``requires an agentive subject'').  We show in section
\ref{E-context} below the role played by metonymy in argumentation.

\begin{el}
\ex
\begin{eli}
\exi {\em Vehicle B \underline{seemd to want} to let vehicle A go
through,\/} (B42)
\label{1}

\exi {\em Being momentarily stopped in the right lane on Boulevard des
Italiens, I had switched my blinker on; I was at a stop and getting
ready to change lanes.  Vehicle B coming from my left
\underline{squeezed} too close to me and damaged the whole left
front side.\/} (A7)\label{A7}
\end{eli}
\end{el}

\begin{el}
\ex
\begin{eli}
\exi {\em my bumper\/} (A11)\label{bumper}

\exi {\em Je roulais\/} ({\em I was driving\/}, literally {\em I was
rolling\/})
\label{rolling}
\end{eli}
\end{el}

\subsubsection{Spatial and physical knowledge}

As part of K-context, the knowledge of a number of spatial or physical facts
(cinematics, dynamics, etc.)  is required to understand what happens in a
text about road accidents.  For example, in \ref{A3}, {\em R\/} must be able
to conclude that vehicle B is to the right of vehicle A.  The reasoning is
obvious:  swerving to avoid a car door on the left can only be done with any
plausibility towards the right.  Nevertheless, for an automatic treatment of
such inferences, we must give all the rules needed for reconstructing
this natural reasoning.  \cite{Gayral92} and \cite{Gayral/etal94} attempt to
provide such a model, based on naive physics.

In the case of text \ref{A8}, {\em R\/} will deduce without any difficulty
that vehicle B was driving too fast when it entered the curve and that this
excessive speed was the cause of it being thrown off course.  But this
excessive speed is only mentioned in the second sentence of the text, and
there is no linguistic motivation for associating it with the previous event
{\em arriving in front of me in the curve\/} described in the first
sentence.  However, this is what K-context allows {\em R\/} to do.  On
the one hand, when a vehicle is {\em thrown off course\/}, some typical
reasons such as a slippery road, high speed, or a mechanical incident come
to mind; these can include a curve linked with high speed.  On the other
hand, when a vehicle is ``thrown off course in a curve'' one usually does
not speed up but rather brakes.  So, in \ref{A8}, since the vehicle had an
excessive speed after being thrown off course, it would necessarily have had
this high speed before being thrown off course.

\subsection{F-context}
\label{F-context}

\subsubsection{Parameter ``Accident''}

In many of our texts, the accident is explicitly mentioned with verbs such
as {\em percuter, endommager, toucher, heurter\/} (``collide'', ``damage'',
``touch'', ``hit''), or with nouns such as {\em choc, collision\/}
(``impact'', ``collision'').
Interestingly, the word {\em accident\/} itself almost never occurs, and the
accident is evoked with some more or less complex paraphrase.  This
aspect of the texts is linked to E-context and the argumentative component in
an accident report.  The writers use circumlocutions to emphasize the idea
that the accident happened {\bf in spite of} all their efforts to avoid it.
One of the best examples is \ref{certainement}:

\begin{el}
\ex {\em We certainly got closer and consequently hit each other, her car
getting stuck into mine, its left fender into the right front side of my
car.\/} (A17)\label{certainement}
\end{el}
But this is not the case in \ref{A15}, a text for which, if it was another
type of narrative, we might imagine other endings to the incident (e.g.\
{\em but I was able to swerve and avoid it\/}).

\begin{el}
\ex {\em We were in Saint-Ouen, I was surprised by the person who
braked in front of me, not being able to change lanes, and the road
being wet, I couldn't stop completely in time\/} (A15)\label{A15}
\end{el}

We can see here the effect of the ``Accident'' parameter:  since these texts
are accident reports, the series of events they relate must by default
contain an accident.  The interpretation of the texts often requires the
reconstruction of an impact between the two vehicles, as in \ref{A15},
where the incident which is described would not otherwise warrant the
existence of the report.  We will see more instance of this in section
\ref{E-context}, where we look at its argumentative effect.

The existence of the impact can then be deduced from a combination of
several clues, some linguistic, some inferential.  Among the former,
we often find the combination of the negation with a verbal group of
the form ``can/be able to $+$ V'', for instance {\em I couldn't stop
completely in time\/} in \ref{A15}, or {\em I wasn't able to avoid the
car which was coming with great speed\/} in \ref{A8}.

\subsubsection{Parameter ``Participants''}

The fact that car accidents usually involve two participants, most often two
vehicles, is used to infer the identity of some entities in the texts or to
establish coreference between two entities.

A specific naming convention in French insurance claim reports for
the vehicles involved in an accident is that claimants must refer to their own
vehicle and to their opponent's as A or B.  This convention arises
from the pre-defined format of the claim report, on which each of the two
drivers must first answer a set of questions by checking boxes in one of two
columns A or B, thus choosing for themselves one of the roles.
They usually then continue to use these labels for themselves and their
opponent in the free-running text, but not necessarily in the whole text.
Indeed the authors often mix first person expressions with these
neutral third person labels.

{}From {\em R\/}'s point of view, resolving the problem of reference,
i.e.\ identifying the different vehicles involved in the accident and their
drivers, often requires knowledge of F-context, particularly of the
``Participants'' parameter.  We now look at several examples where knowing
that there are two vehicles involved in the scene of the accident helps
resolve anaphors.

\begin{el}
\ex {\em I was going down towards Bellefontaine.  The road is a narrow,
windy road, lined with trees.  In a curve with not much visibility,
\underline{we} collided.\/}  (B33)\label{B33}
\end{el}

The pronoun {\em we\/} in \ref{B33} refers to the two vehicles involved in
the accident, although the opponent's vehicle is not mentioned (the writer's
vehicle is implicit in {\em I was going down\/}).  This anaphor can only be
resolved because of the F-context ``Participants'' parameter.  Without the
context that there should be two vehicles involved, the pronoun {\em we\/}
would be surprising and probably uninterpretable.

\begin{el}
\ex
{\em \underline{Vehicle A} waiting and stopped at the Pont de
Levallois lights.  Vehicle B arrived and hit \underline{my} left side
mirror with its right side mirror.\/} (C10)\label{A-B}
\end{el}

The first proposition in \ref{A-B} introduces vehicule A, i.e.\ one of the
two vehicules involved in the accident.  The second proposition introduces
vehicule B, i.e.\ the other vehicule involved in the accident.  In the third
proposition, the reader encounters {\em my left side mirror\/}.\footnote
{The use of the metonymy in the expression {\em my left side mirror\/}
reinforces the sense of hesitation about reference which we observe in our
texts.}
If {\em R\/} did not know the convention, this expression would force the
introduction of a third vehicle, which would have to be {\em W\/}'s because
of the {\em my\/}.  Indeed, without the knwoledge that {\em W\/}'s vehicle
is named A or B, there is no reason to identify vehicle A with it,
even if vehicle A's role in the scene is then rather unclear.  However, with
the knowledge of this convention, coreference can be resolved.

Another example is \ref{B28}, where vehicle B is mentioned in the first
sentence and the second vehicle involved in the accident is mentioned in
the second sentence.

\begin{el}
\ex
{\em Coming back home, the driver of vehicle B in front of me lost control
of his vehicle because of sudden icing.  In turn I couldn't control my
vehicle which after 20 meters crashed into Mrs.\ Louvet's vehicle.  I want
to stress that there was no ice anywhere else and we were many vehicles
skidding on this street.  Nothing could allow foreseeing such icing
conditions.\/} (B28)\label{B28}
\end{el}
The relative clause in the second sentence refers to an accident between
Mrs.\ Louvet's vehicle and {\em W\/}'s.  Nothing, except F-context, warrants
linking vehicle B and Mrs.\ Louvet's, and at first glance, there could
appear to be three vehicles in this scene.  However, because of the
F-context ``Participants'' parameter, and because the text says that the
accident takes place between {\em W\/}'s vehicle and vehicle B, the reader
can deduce that Mrs.\ Louvet is the driver of vehicle B and that there are
only two vehicles involved.

\subsection{E-context}
\label{E-context}

When setting to the task of writing such a report, {\em W\/} knows
the ``Short'' parameter, the constraint that only about a paragraph (in a
pre-defined area on the form) may be used to relate the accident.  At the
same time, {\em W\/} must not forget any important information whose absence
would prevent {\em R\/} from reconstructing the correct factual content {\em
P\/}, and he must thus be both {\bf exhaustive} and {\bf concise}.

On the other hand, the authors know that these few lines, meant for their
insurance company, may contribute to the final decision about their share of
legal and financial liability.  They know that the intended readers, the
insurance agents, must pass a judgement on their behavior and will determine
their share of responsibility in the accident.  Necessarily then, the
authors of those reports attempt to present their case in the best possible
light in order to minimize their responsability.

In short, {\em W\/} is faced with what we call ``{\em W\/}'s selection
problem'', namely the constraint on the choice of information to
give in order to satisfy the three goals:  to be exhaustive, to be concise,
and to lessen their responsability.\footnote{
This problem is part of the wider language conventions LC and constitutes a
``meta-knowledge'', essential for the success of communication:  the text
{\em T\/} must provide all the information that is necessary in order to be
understood or to convince, but only that much.  This problem can be
considered a particular instance of Grice's Maxims \cite{Grice75}, in
particular the Maxim of Quantity, or of Ducrot's exhaustivity law
\cite{Ducrot72}.}

These goals are not contradictory and actually become intermingled.  While
describing the scene, {\em W\/} is trying to argue for his innocence.  Thus,
the choices of which elements are mentionned in the report can thus reveal
an argumentative strategy while helping reconstruct the factual content of
the text.  The authors can choose to adopt a ``legal'' framework for
describing the setting of the accident.  They then try to speak the same
language as the insurance agent and give exactly the information that the
latter expects.  They choose precise words to refer accurately to the
objects which are present in this space seen from a legal point of view and
which are directly relevant to traffic, e.g.\ road signs, markings, or
referring to events happening in this domain also from a legal point of
view, such as {\em turn his blinker on\/}, {\em coming from the left\/},
etc.

Since it invokes legal traffic rules, this information also evokes some
particular behavior on the part of the drivers involved.  Most of the time,
however, the presentation given by {\em W\/} is not neutral but aims to
prove either that his behavior was the one expected in the situation, or
that his opponent's behavior was wrong and unexpected, or even both at the
same time.  The argumentation may be explicit or left implicit.  For
instance, in \ref{A12-markings}, the first two sentences set up in a very
detailed way a situation which is specifically controlled by traffic rules.

\begin{el}
\ex {\em I was driving in my vehicle A in the right lane reserved for
vehicles going straight ahead.  Vehicle B was driving in the left lane
reserved for vehicles going left (ground markings with arrows).  It
cut back in on my vehicle.\/} (A12-markings)\label{A12-markings}
\end{el}

The presence of ``ground markings with arrows'' implies a particular
behavior on the part of vehicles driving in that lane:  they must go left.
The last sentence exactly contradicts this expected behavior and is intended
to prove that vehicle B made a mistake in ``cutting back in'' on {\em W\/}'s
vehicle since it was not respecting the ground markings.  Of course, {\em
W\/} has taken care to mention that his own vehicle was in the correct lane.

In \ref{right-of-way}, {\em W\/} clearly refers to an important French
traffic rule, namely that the right-of-way always belongs to the vehicle
coming from the right.  This right-of-way is implied in the first sentence:
vehicle B arrived from the left at an intersection and should have let
vehicule A go.  Since an accident has occurred, {\em R\/} may then deduce
B's wrong behavior, which {\em W\/} then refers to explicitly at the end of
the text.  At the same time, {\em W\/} takes care to insist on his own
correct behavior ({\em at moderate speed\/}).

\begin{el}
\ex {\em Vehicle B coming from my left, I find myself at the
intersection,  \underline{at moderate speed}, about 40 km/h, when vehicle B
hits my vehicle, and \underline{denies} me the \underline{right-of-way} from
the right.\/} (A4)\label{right-of-way}
\end{el}

In \ref{A11}, the first sentence indicates that the driver of vehicle B did
something illegal, since passing a vehicle must, according to French traffic
rules, be done on the left.  The end of the text then reinforces B's wrong
behavior through the use of lexical elements such as {\em slalom\/} and {\em
ran away\/}.

\begin{el}
\ex {\em The driver of vehicle B passing me on the right caught my right
front bumper and dragged me towards the movable wall on the Genenevilliers
Bridge, which I violently smashed into.  According to the witness who was
following me, the driver of vehicle B was \underline{doing a slalom} between
the cars.  After hitting me, he ran away and couldn't be caught up with by
the above-mentioned witness.\/}  (A11)\label{A11}
\end{el}

The authors often express their own psychological states (e.g.\ {\em
\^{e}tre surpris\/} ``to be surprised'' in \ref{surprise}) or thoughts
during the accident ({\em je ne m'attendais pas\/} ``I didn't expect'' in
\ref{not-expect}), although these are not a priori directly interesting for
the insurance company.

\begin{el}
\ex {\em I was \underline{surprised} by the person who braked in
front of me, \underline{not being able to} change lanes, and the
\underline{road being wet}\/} (A15)\label{surprise}
\end{el}

\begin{el}
\ex {\em I \underline{didn't expect} that a driver would wish to pass
me for there weren't two lanes marked on the portion of the road where
I was stopped.}  (A5)\label{not-expect}
\end{el}

Mentioning them allows {\em W\/} to explain his behavior, particularily in
establishing a contrast between what was expected and what happened in
reality.  {\em W\/} can claim not to have had any control over what was
actually happening because that was a consequence of unforeseeable and/or
uncontrollable circumstances:  the road was wet in \ref{wet}, the pavement
was slippery in \ref{slippery}.  Thus {\em W\/} cannot be considered as
responsible for the accident.

\begin{el}
\ex
\begin{eli}
\exi {\em on impact, and because of the slippery pavement, my vehicle skids,
and hits the metal railing around a tree, whence a second front impact.\/}
(A4)\label{slippery}

\exi{\em and the road being wet, \underline{I wasn't able} to stop
completely in time.} (A15)\label{wet}
\end{eli}
\end{el}

The very frequent use of negation is also a favorite clue to indicate
implicitly an opposition between what should have happened and what actually
occurred, for instance in in \ref{wet} and in \ref{not-able}.  The use
of negation can also be a way of not mentioning explicitly the collision
(see section \ref{F-context}).

\begin{el}
\ex {\em I \underline{wasn't able} to avoid the car which was coming
with great speed.\/} (A8)\label{not-able}
\end{el}

Another argumentative device is the reverse of the metonymy conflating the
vehicle and its driver which we saw in \ref{K-context}.  For instance, in
\ref{slippery}, it is not {\em W\/}, but the car which is the subject of the
two verbs, as if it was responsible for the events.  Because they suppress
the agent, reflexive verbs ({\em la porte s'est ouverte\/} ``the door
opened'') or the passive voice ({\em j'ai \'{e}t\'{e} d\'{e}port\'{e}\/} ``I
was thrown off course'') instead of a plain active, are two
contructions which also help suggest that {\em W\/} was not involved in the
course of events and cannot be held responsible for what happened.

To summarize, the examination of the terms used and of the elements which
the authors choose to mention in their texts reveals that there are two
strategies they can follow to argue their case in the most persuasive way:

$\bullet$ A.  Trying to push the blame onto the opponent by accusing him
of abnormal behavior

$\bullet$ B.  Contrasting what was expected and what happened in reality,
by invoking unforeseable circumstances.

With either strategy, {\em W\/} must first show that he has done everything
that was required in the given circumstances and will always try to appear
as blameless as possible.  Of course the two strategies are not mutually
exclusive as shown by \ref{A15} above.  \ref{A14} is also an example of a
mixture of both strategies, in which where {\em W\/} piles up all sorts of
attenuating circumstances and also emphasizes ({\em immediately put the
brakes on\/}) his own appropriate reactions.

\begin{el}
\ex
{\em I was driving at about 45 km/h in a \underline{small one-way}
street where cars were \underline{parked on both sides}.  Popping
\underline{suddenly} on my right coming out of a private building
garage, Mrs.\ Glorieux's vehicle was at a \underline{very short
distance} from my vehicle; passage being \underline{impossible}:
\underline{surprised}, I immediately put the brakes on but the impact
was unavoidable.}  (A14)\label{A14}
\end{el}

\section{Inferences}
\label{inferences}

In this section, we give evidence for the role which the knowledge of the
argumentative function of such texts plays in the process of interpretation,
particularly in the reconstruction of its factual content.

\subsection{Lexical Ambiguity}

As shown in \ref{A8.rep}, the original French text of the example given in
\ref{A8} presents a an exemple of this kind of ambiguity, since in French,
the word {\em droite\/} is ambiguous between the two interpretations {\em
right\/} and {\em straight\/}.\footnote{ If the adjective {\em droite\/}
means {\em straight}, its opposite is then {\em courbe/curved\/}, if it
means {\em right\/}, the opposite is then {\em gauche/left\/}.}

\begin{el}
\ex
Je roulais sur la \underline{partie droite} de la chauss\'{e}e (A8)

{\em I was driving on the \underline{right-hand side / straight portion} of the
road}
\label{A8.rep}
\end{el}

Here, even though the whole text can also be interpreted with the {\em
straight\/} meaning, the {\em right\/} interpretation is more plausible.
However, only an argumentative type of reasoning can lead {\em R\/} to
prefer the latter.  Since it is well-known to both {\em R\/} and {\em W\/}
that in France one drives on the right, by specifying that he was driving on
the {\em right\/} side of the road, {\em W\/} violates the Maxim of Quantity
(i.e.\ not to say anything superfluous) and therefore must be taken as
intending to convey some other information.  In this case, it must be to
assert that his behavior was conforming to the {\em Code de la route\/} (the
``Rules of the Road''), which is indeed a pertinent fact to mention.  Here,
informational redundancy by itself carries some information which allows
inference.  We can thus formulate the rule that:  ``In case of ambiguity,
{\em R\/} should prefer the interpretation from which correct behavior on
{\em W\/}'s part can be inferred''.

\subsubsection{Time Reference Ambiguity}

In the first sentence of \ref{A7}, given in \ref{A7-ppf}, the use of the
pluperfect {\em had switched on\/} is ambiguous.

\begin{el}
\ex {\em Being momentarily stopped in the right lane on Boulevard des
Italiens, \underline{I had switched my blinker on};  I was at a stop and
getting
ready to change lanes.\/} (A7)\label{A7-ppf}
\end{el}

The pluperfect implies that the process being talked about is perceived with
another past event as a point of reference, which may not yet have been
mentioned.\footnote{
The situation is exactly parallel in French and English.}
Here, two different referential situations can be envisaged, with two
different consequences:

$\bullet$ If the accident itself is chosen as the point of reference,
switching the blinker on signals a future change of lanes.  It must
therefore be the left blinker.  This conclusion requires geometrical
reasoning:  ``If X is stopped in the right lane and if X wants to
change lanes, X can only go left''.

$\bullet$ If the time of stopping is chosen as the point of reference,
switching the blinker on is prior to the time of stopping and thus
signals it.  It must then be the right blinker, since the vehicle is
in the right lane.

Arguments of the ``Maxims'' type must then be used.  {\em R\/} cannot assume
that too much information is present in the text.  For the blinker to be
switched on before stopping would not be relevant since the accident
occurred after that of stopping, when {\em W\/} started again.  On the other
hand, the fact that {\em W\/} did switch his blinker on before starting
again is very relevant from an argumentative point of view, since it means
``{\em W\/} behaved in the right way and did what was required''.
Therefore, by the rule proposed above, the first interpretation is chosen
and {\em R\/} may conclude that {\em W\/} had his left blinker on.

\subsubsection{Action or Intention?}

Sometimes, the problem for {\em R\/} is to determine whether an action
presented as an intended future event has remained at a purely
intentional level or whether actions have already been taken to reach it.
For instance, when the intended action belongs to a script with
sequential steps, the question arises whether some of the preparatory
actions belonging to the script have already been accomplished.

We have seen that there are two possible choices for a point of reference in
the interpretation of the pluperfect in \ref{A7-ppf}.  In addition, the
verb {\em s'appr\^{e}ter \`{a}\/} can have several interpretations.  Like
{\em to get ready\/} (which we give here as its translation), it can mean
{\em to be about to\/} and then it is a simple aspectual auxiliary focussing
on the beginning of the action (inchoative).  It can also have a more
agentive interpretation and then it means {\em to actively prepare for\/}.

In the inchoative {\em to be about to\/} interpretation, the action of
``switching the blinker on'' is an event independent of ``changing
lanes''; in the agentive {\em to prepare for\/} interpretation, that
same action corresponds to one of the preparatory acts.  But more
crucially, in the agentive interpretation, {\em W\/} may already have
started changing lanes and then probably would be at fault, while in
the inchoative reading, {\em W\/} would still be stopped and would be
innocent.

It seems that in most of the cases we find in our texts, such an intended
future event is more than simply intentional and that {\em W\/} has indeed
already started to act.  Otherwise it would not be possible to explain the
accident in \ref{A7-ppf} (\ref{A7}), since there would be no reason for {\em
W\/}'s car to have been damaged if {\em W\/} had not already started turning
left.

Similarly in the case of the texts given in \ref{A2} and
\ref{A5} below, the only plausible reconstruction of the accident requires
vehicle A to have already started the action which is presented as an
intention ({\em Wanting to pass a hauler\/} in \ref{A2} and {\em I wanted
to enter the second lane\/} in \ref{A5}).

\begin{el}
\ex {\em \underline{Wanting to pass a hauler} with its right blinker
on, the latter turned left, forcing me to steer left to avoid it.  The
car skidded on the wet pavement and struck a sidewalk then a fence
straight ahead.  The truck driver had indeed switched on his left
blinker, but the trailer was inverting the signal to the right.  Not
having touched me, the driver declared himself unconcerned by the
situation and refused to draw a report.  Having left my car to call a
mechanic, I came back to find it with the right back door bashed in
with no note left by the guilty party.}  (A2)
\label{A2}
\end{el}

\begin{el}
\ex {\em I was stopped at the intersection wishing to take the road on
which the intense traffic is going one-way in two lanes; as the last
vehicle of the flow was coming, \underline{I wanted to enter}
\underline{the second lane}, leaving the first one free for it.  The
moment I started, I heard the shock in the back; I wasn't expecting a
driver would wish to pass me for there weren't two lanes marked on the
portion of the road where I was stopped.}  (A5)\label{A5}
\end{el}

Instead of using an imperfective verbal form (i.e.\ {\em \'{e}tant en
train de d\'{e}passer un semi-remorque\/} (``while passing a hauler'')
in \ref{A2}, or {\em j'\'{e}tais en train de tourner \`{a} gauche\/}
(``I was turning left'') in \ref{A5}) which would clearly indicate
that the action had already started, {\em W\/} chooses the intentional
form and in doing so, creates an ambiguity for {\em R\/}:  ``Had {\em
W\/} actually already done something or not?''.  This lack of
precision (or downright lie?)  is intentional and allows {\em W\/} to
try to lessen his responsibility.  This will succeed if {\em R\/} opts
for a purely intentional reading of the verbal form.

\subsubsection{Argumentation-Based Inference}
\label{argument-inference}

We said earlier that in \ref{A1} {\em R\/} could infer that,
though this is not stated in the text, {\em W\/} probably wanted to turn
right, but that discourse argumentation was required for this conclusion.
Indeed, the script for going through stop-sign says that ``If X wants to
turn right at a stop-sign, X should check to the left; if X wants to
turn left, X should check to the left and to the right.''.

Since {\em W\/} does not mention checking to the right but only checking to
the left, it means that {\em W\/} did not intend to turn left.  If {\em W\/}
was going to turn left, to mention checking to the right would be pertinent
information for the insurance company since it would show that {\em W\/} had
done everything required in such circumstances.  In fact, it is the
non-homogeneity of this discourse which suggests that {\em W\/} did turn
left:  either no check should be mentioned or if only one is mentioned, then
{\em R\/} may infer that the other one was not required.  In another line of
argument, checking to the left is also mentioned by {\em W\/} in order to
explain that, since he was looking to the left (and not straight ahead), he
could not have seen that the other car had not turned right.

\section{Conclusion}

In this paper, we address the general issue of ``how to define context'' and
we have chosen an experimental rather than a theoretical approach to this
question.  By selecting real occurring texts, instead of texts written to
illustrate particular phenomena, and an homogeneous corpus of texts written
in similar circumstances, we were able to focus on the characteristic
properties of this text type and thus to better define the notion of
context.

We have tried to show the importance of situational, cultural and textual
presuppositions from the point of view of both the writer and the reader.
As this work constitutes a first step in the study of natural language
semantics in the context of an NLP project, the approach adopted here is an
attempt to automate the process of understanding these texts and deriving
inferences from them.  Some of the crucial issues in NLP are precisely how
to define and describe the different types of knowledge involved in the
processes of writing and reading texts, and how to establish rules which
mimic the reasoning involved in these activities.

Here, we take advantage of the specificity of the texts-- the authors
narrate events leading to a car accident while trying to lessen their
responsability-- to circumscribe the type of knowledge required and to give
some rules of interpretation, valid for this type of text, in this type of
context.

We have determined three types of context:  K-context, the non-linguistic
knowledge required for this domain, F-context, the more specific context of
the events being narrated, and E-context, the discourse context for this
text type.  The interest of the corpus we have chosen lies in the fact that
it contains texts which involve the same three types of contexts:  K-context
because they all deal with road traffic, F-context because they all deal
with car accidents and E-context because they are all insurance claim
reports.

We have shown the importance of E-context, in particular the crucial role
played by the argumentation which the writer is known to be pursuing and
which allows the reader to make a number of inferences.  These inferences
then help him clarify the text and choose between competing interpretations.

It would be interesting to analyze the two corresponding texts, by
two opponents reporting the same accident, in order to establish which
part of the information is objectively factual and shared by both
texts, and which part of the information is argumentatively biased,
thus better distinguishing the subjective part of the two discourses.
The omission of information, which we mentioned as one of the
argumentative devices on the part of {\em W\/} and as a basis for
inference on the part of {\em R\/}, would then become an even more
important factor in the analysis.  Very few such pairs of texts are
available, but in the continuation of this project, we may try to do
some further work based on those we have.

\end{document}